\documentclass[sigconf]{acmart}
\AtBeginDocument{%
  }

\setcopyright{acmlicensed}
\copyrightyear{2025}
\acmYear{2025}
\setcopyright{licensedusgovmixed}\acmConference[SIGSPATIAL '25]{The 33rd ACM International Conference on Advances in Geographic Information Systems}{November, 2025}{Minneapolis, MN, USA}
\acmBooktitle{The 33rd ACM International Conference on Advances in Geographic Information Systems, November, 2025, Minneapolis, MN, USA}
\acmISBN{978-1-4503-XXXX-X/2018/06}

\usepackage{algorithm}
\usepackage{multirow}
\usepackage{soul}
\usepackage{graphicx}
\usepackage{subcaption}
\usepackage{algpseudocode}
\usepackage{todonotes}
\usepackage{makecell}
\usepackage{pgfplots}
\usepgfplotslibrary{groupplots}
\pgfplotsset{compat=newest}

\usepackage{colortbl}
\begin{document}

\title[Geospatial Diffusion for Land Cover Imperviousness Change Forecasting]{Geospatial Diffusion for Land Cover \\ Imperviousness Change Forecasting}


\author{Debvrat Varshney}
\authornote{Both authors contributed equally to this research.}
\orcid{0000-0001-8898-1736}
\affiliation{%
  \institution{Oak Ridge National Laboratory}
  \city{Oak Ridge}
  \state{TN}
  \country{USA}
}
\email{varshneyd@ornl.gov}

\author{Vibhas Vats}
\authornotemark[1]
\orcid{0000-0002-7232-0121}
\affiliation{%
  \institution{Indiana University Bloomington}
  \city{Bloomington}
  \state{IN}
  \country{USA}
}
\email{vkvats@iu.edu}

\author{Bhartendu Pandey}
\orcid{0000-0002-3712-5961}
\affiliation{%
  \institution{Oak Ridge National Laboratory}
  \city{Oak Ridge}
  \state{TN}
  \country{USA}
}
\email{pandeyb1@ornl.gov}

\author{Christa Brelsford}
\orcid{0000-0002-3490-8020}
\affiliation{%
 \institution{Los Alamos National Laboratory}
 \city{Los Alamos}
 \state{NM}
 \country{USA}
 }
\email{cbrelsford@lanl.gov}

\author{Philipe Dias}
\orcid{0000-0001-9427-7112}
\authornote{Corresponding author}
\affiliation{%
  \institution{Oak Ridge National Laboratory}
  \city{Oak Ridge}
  \state{TN}
  \country{USA}
}
\email{ambroziodiap@ornl.gov}


\begin{abstract}

Land cover, both present and future, has a significant effect on several important Earth system processes.  For example, the Urban Heat Island effect, driven in large part by impervious surfaces, can increase urban land surface temperatures by up to 5 \textdegree C. Impervious surfaces heat up and speed up surface water runoff and reduce groundwater infiltration, with concomitant effects on regional hydrology and flood risk. While regional Earth System models have increasing skill at forecasting hydrologic and atmospheric processes at high resolution in future climate scenarios, our ability to forecast land-use and land-cover change (LULC), a critical input to risk and consequences assessment for these scenarios, has lagged behind.    


In this paper, we propose a new paradigm exploiting Generative AI (GenAI) for land cover change forecasting by framing LULC forecasting as a data synthesis problem conditioned on historical and auxiliary data-sources. We discuss desirable properties of generative models that fundament our research premise, and demonstrate the feasibility of our methodology through experiments on imperviousness forecasting using historical data covering the entire conterminous United States. Specifically, we train a diffusion model for decadal forecasting of imperviousness and compare its performance to a baseline that assumes no change at all. Evaluation across 12 metropolitan areas for a year held-out during training indicate that for average resolutions $\geq 0.7\times0.7km^2$ our model yields MAE lower than such a baseline. This finding corroborates that such a generative model can capture spatiotemporal patterns from historical data that are significant for projecting future change. Finally, we discuss future research to incorporate auxiliary information on physical properties about the Earth, as well as supporting simulation of different scenarios by means of driver variables.


\end{abstract}



\received{20 February 2007}
\received[revised]{12 March 2009}
\received[accepted]{5 June 2009}


\maketitle


\section{Introduction}\label{sec:intro}


Characterization of land-use and land-cover (LULC) has been a major research topic in the context of geospatial data analysis as it has widespread relevance for measuring urbanization, understanding local context in data sparse regions, and in performing risk and consequence assessment for Earth and environmental hazards. 
For example, the Urban
Heat Island effect, driven in large part by impervious surfaces, can involve increases in urban land surface temperatures by up to 5 °C \cite{SHI2023101529}. Impervious surfaces can also speed up surface water runoff and reduce groundwater infiltration, with concomitant effects on regional hydrology and flood risk \cite{COON2022105502,feng2021urbanization,MIGNOT2019334}.  



Advancements in remote sensing and deep learning have enabled substantial progress on scalable and high-quality characterization of historical LULC. Datasets such as USGS's National Land-Cover Database (NLCD) maps the conterminous United States (CONUS) at $30m/px$ \cite{dewitz2023} using Landsat data, while ESA's World Cover data \cite{zanaga2022esa}, Esri's Land Cover \cite{karra2021globalESRI}, and the Google's Dynamic World \cite{brown2022dynamic} all provide global coverage at $10m/px$ based on Sentinel data. However, our ability to model and forecast land cover change still remains an open research problem with ever increasing relevance.



For example, forecasting land cover imperviousness is critical for understanding potential future hydrologic processes. Impervious surfaces cause water to accumulate and run off much more quickly than it does for natural or pervious surfaces such as vegetation, contributing to ``flashier" hydrologic systems. The concentration of rainfall runoff into a smaller time period then significantly increases the risk of flooding. Consequently, our ability to assess such risks for future environmental hazards is significantly impaired, due to limited land cover forecasts. Furthermore, urbanization is expected to increase the impervious surface area to more than twice its current extent over the next 3 decades \cite{seto2012global,gao2020mapping}, making the assumption of no land cover change increasingly untenable.  
LULC change is very difficult to forecast, as it is primarily driven by humans acting individually or collectively in response to changing preferences, attitudes, capabilities, and needs. While behavior is never fully deterministic, there are stochastically observable patterns in land use choices which make forecasting possible. 
For example, SLEUTH \cite{clarke1997self}, one of the earliest land cover change forecasting models, assumes that increases in impervious surface area occur primarily along existing transportation networks. Many existing land cover forecasting strategies adapt similar heuristic approaches for assessing land cover change probabilities.  
 
In this paper, we propose a new strategy of applying Generative AI (GenAI) techniques to the LULC forecasting problem. Inspired by the recent breakthroughs in GenAI for modelling and synthesizing data from complex spatiotemporal distributions (e.g., text-to-image synthesis \cite{dhariwal2021diffusion} and video forecasting \cite{voleti2022mcvd}), we discuss a new spatial modeling paradigm leveraging diffusion models to synthesize empirically plausible forecasts of LULC change. GenAI offers the following key desirable properties that motivate its application for this problem:

\begin{enumerate}
    \item GenAI can capture spatiotemporal patterns of change that go beyond assumptions of spatial connectivity and, in contrast to Markovian approximations, consider multi-temporal associations;
    \item GenAI approaches such as diffusion models can synthesize multiple probabilistic outputs. This is beneficial since LULC forecasting is a non-deterministic problem where the interest is not on a single output, but rather an estimation of the most likely future scenarios;
    \item The data-driven nature of conditional GenAI enables leveraging the increasingly larger datasets of historical LULC and auxiliary geospatial data (e.g., population records, topography, and critical infrastructure), learning potentially non-linear interactions between LULC change drivers;
    \item Compared to conventional region-specific models, a GenAI-based forecasting framework is positioned to better leverage the large datasets currently available to enable high-accuracy forecasts while generalizing across regions. 
\end{enumerate}


In this paper, we demonstrate the feasibility of this approach by introducing a diffusion-based framework for decadal land cover imperviousness forecasting. Trained on NLCD's historical imperviousness and LULC data across the CONUS, we show that this approach outperforms at sub-kilometer level a ``null'' model that assumes no change (a strong baseline since land cover change is heavily imbalanced), generalizing across a diverse selection of evaluation regions. We also provide qualitative and quantitative comparisons of our model's forecasting ability to a more traditional and popular CA-Markov based method for land cover forecasting. 


\section{Background and Related works} 
\label{sec:related-works}



\noindent\textbf{Traditional approaches for LULC change forecasting.} Urbanization is complex, with existing tools for projecting land cover change being limited. Methods for estimating urban growth patterns usually assume urban homogeneity and hence are region-specific \cite{zhou2019highSLEUTH}. FORE-SCE was a method developed to project LULC change under a variety of climate and socioeconomic scenarios \cite{Sohl26062007}, but except for the information contained within historic LULC products, this model lacked information on urbanization and urban growth. 

Methods such as Cellular Automata (CA) models have been a major strategy used to study land use change. A CA model, in its simplest form, consists of a discrete definition of a region of cells, paired with a set of transition rules based on the region's own state and the state of its' neighboring cells. A \textit{CA Markov} model adds Markov chain analysis to a basic CA model, allowing probabilistic transitions between land cover states.  

SLEUTH is a widely used CA model formulated with a small number of decision rules and parameters to forecast urban growth and land-use change \cite{clarke2007decade, zhou2019highSLEUTH}, but has limited capacity to represent complex spatial growth phenomena. While CA models have demonstrated usefulness for land cover modeling and forecasting, they face two key challenges. First, transitions are based on local information only, and while many interesting structures can be generated from local interactions, urban environments in particular are likely to be influenced by both local context and broader regional context \cite{seto2012urban}. Second, its land cover transition rules must be defined a priori, not enabling the model to learn them from the data. Given the complexity of human behavior and choices around land cover change, this can be a limitation.

\noindent\textbf{Deep learning based approaches.} There have been a variety of works using popular deep learning-based approaches for this problem domain. Methods such as \citet{metzger2023urban} forecast new building constructions from satellite imagery, by using a U-Net pretrained for building change detection and fine-tuning it to classify pixels into early-change (1-12 months) or late change (12-24 months). Their model obtained F1 scores of around $15\%$ for 24-months prediction range, anticipating new buildings at existing construction sites and some sprawl to wastelands in their vicinity. The authors highlighted that making such forecasts from a single image is an ill-posed and very challenging task. Further, Recurrent Neural Networks (RNNs) and Long-short Term Memory (LSTMs) have also been explored for simulating urban land dynamics. \citet{liu2021simulation} employed LSTMs to augment CA-based forecasting with transformation rules that better capture long-range time dynamics while \citet{zhu2024simulating} introduce cycle-consistent learning scheme to RNNs for forecasting LULC changes.





\noindent\textbf{Background on diffusion models.} 
AI models tackling tasks such as image LULC classification are typically \textit{discriminative models}, which focus on learning boundaries separating data points into categories. In contrast, generative AI targets modeling the data distribution such that realistic samples can be synthesized. This is a significantly harder task, for which significant progress has been recently achieved by methods including Variational Autoencoders \cite{ramesh2021zero}, Generative Adversarial Networks (GANs) \cite{NIPS2014_f033ed80}, and more recently, diffusion models.  

\begin{figure*}[t]
\begin{center}
    \includegraphics[width=\textwidth]{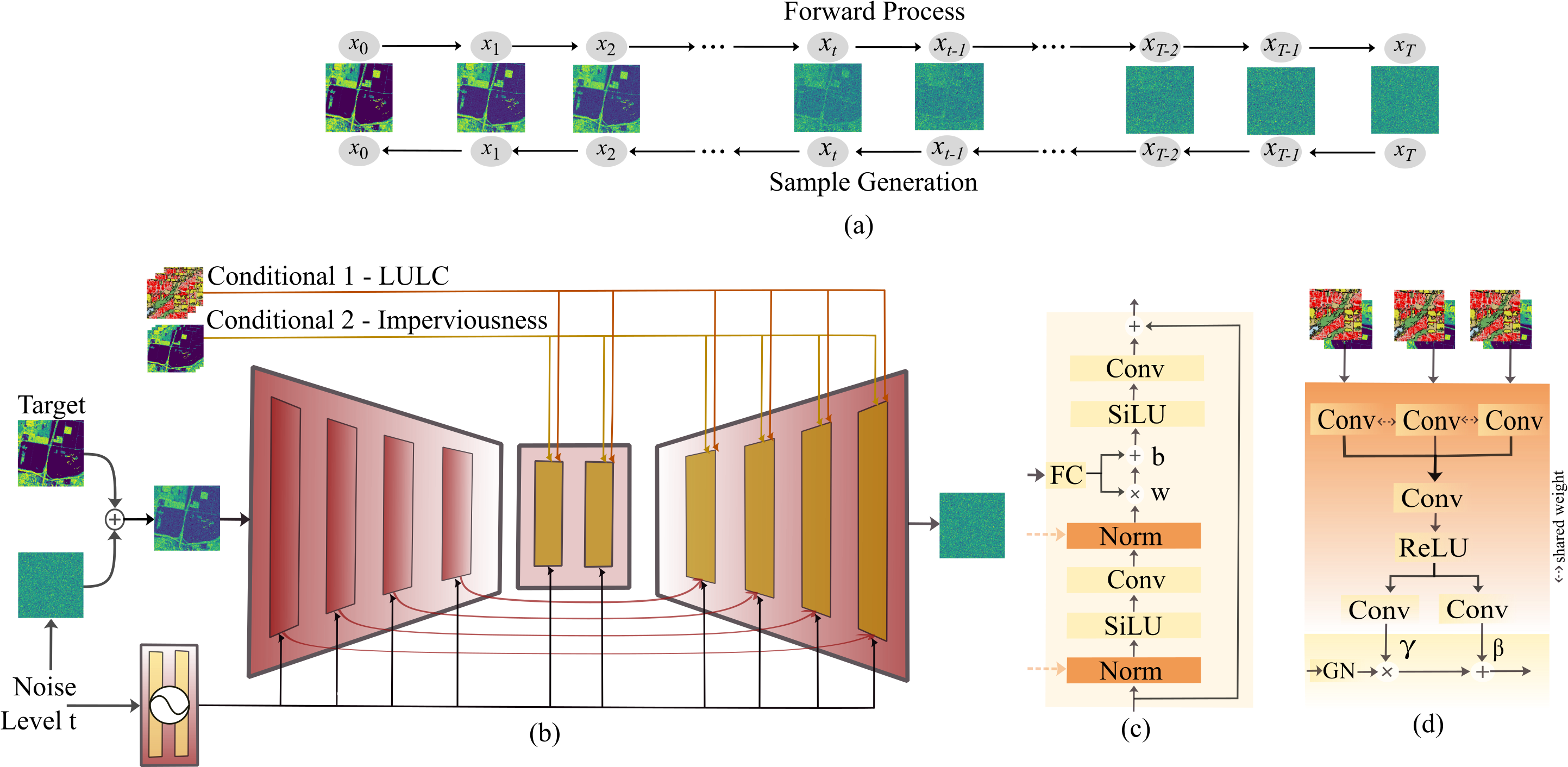}
    \caption{Overview of the diffusion process and the conditional CNN architecture. (a) The forward process involves adding a timestep-dependent noise to the input image, while the 
    sampling process takes place through a reverse process starting at $x_T$. (b) he architecture of the conditional UNet 
    \cite{olaf2015unet} trained to estimate the added noise. (c) The internal design of 
    the residual blocks used in the UNet. (d) Our double conditional normalization 
    layer, used as \textit{Norm} in the residual blocks show in c).}
    \label{fig:diffusion_diagrams}
\end{center}
\end{figure*}

Diffusion models are a class of likelihood-based generative models that adopt a Markov chain approach involving two sub-processes. As illustrated in Fig. \ref{fig:diffusion_diagrams}a), in the \textit{forward process} noise is gradually added to the original data such that its complex distribution is converted into a more tractable one (e.g., a Gaussian distribution). A model is then trained to perform the \textit{reverse process}, i.e., estimate the added noise at each step of the diffusion process, such that through iterative denoising it can synthesize back samples that follow the original data distribution. 
Following the seminal denoising diffusion probabilistic model (DDPM) \cite{ho2020denoising} work, such models power the vast majority of modern image/video synthesis tools, such as the DALL-E and Stable Diffusion models building upon the idea of latent diffusion \cite{Rombach_2022_CVPR, dhariwal2021diffusion}.

In contrast to GANs, diffusion models are able to generate very high-quality and diverse data samples without exhibiting mode-collapse and the training instabilities of GANs. Although the relative slow speed of the reverse process required for model inference represents a drawback, for our application the advantages outweigh the limitations since real-time performance is not a requirement. By leveraging denoising diffusion implicit models (DDIMs) \cite{song2021denoising}, the process of high-quality sample synthesis can take place more than $10\times$ faster than conventional DDPMs.


\noindent\textbf{Generative AI for LULC characterization.}
Multiple works have been exploring GANs for LULC characterization. 
\citet{albert2019spatialGAN} develop a GAN in an image-to-image translation scheme for predicting spatial distribution of built environment based on water masks, population density maps, and nighttime luminosity. 
Fan et al. \cite{FAN2024104093} was the first work to use a DDPM for LULC characterization, introducing a multihead cross-attention fusion module between the DDPM and a vision Transformer \cite{dosovitskiy2021an} to improve LULC segmentation. Shi et al. \cite{rs16234573} similarly combines a segmentation network with a DDPM decoder. Both these works make categorical predictions of land cover classes from imagery, in contrast to the forecasting problem targeted by our work.


\noindent\textbf{Conditioning generative models.} The development of conditioning mechanisms to guide the diffusion process is another critical piece enabling the practical success of modern generative models. Conditional information can come in various forms, such as image synthesis from text captions \cite{ramesh2021zero}, semantic layout \cite{park2019semantic}, image style transfer \cite{isola2017image}. In contrast to methods directly concatenating conditional layouts \cite{isola2017image, ramesh2021zero} as additional inputs to deep networks, \cite{park2019semantic} introduces spatially adaptive normalization (SPADE) as a mechanism for conditioning synthesis on semantic layouts. It employs a shallow CNN-based branch that receives conditioning images as inputs (e.g., semantic layouts) and outputs a scaling and a shifting parameter that modulate the normalization layers within the denoising diffusion model. SEAN \cite{zhu2020sean} extended this idea to multiple conditions: it enables region-adaptive normalization based on a style-matrix plus a segmentation layout, doing so by adding a second CNN conditioning branch such that two pairs of scaling and shifting parameters are regressed and combined through learnable weighted average. In Section \ref{subsec:conditional-normalization} we introduce a strategy inspired in SPADE/SEAN to enable conditioning on multiple modalities covering the same spatial area. 


The highly impactful work by \citet{dhariwal2021diffusion} demonstrates diffusion models beating GANs on conditional image synthesis by using adaptive group normalization to incorporate time step and class embedding into each residual block \cite{wu2018group}. Wang et al. \cite{wang2022semantic} modify SPADE normalization to incorporate semantic maps for conditional normalization in the decoder part of the diffusion model to control the image generation process. For video synthesis, masked conditional video diffusion (MCVD) \cite{voleti2022mcvd} adapts SPADE \cite{park2019semantic} into SPAce-TIme-Adaptive Normalization (SPATIN), where group normalization is conditioned on a temporal combination of past and future frames. 


\section{Methodology}\label{sec:methodology}
Our proposed methodology is based on the premise that the problem of LULC change forecasting can be framed as a task of image synthesis conditioned on historical LULC maps and auxiliary data. This allows us to leverage the recent advances on diffusion-based models reviewed above, which have made a significant impact in conditioned image synthesis tasks and present several desirable properties as listed at the end of Section \ref{sec:intro}. In this paper, we focus on the land-cover \textit{imperviousness} change forecasting to validate the proposed paradigm. Compared to LULC forecasting, this represents a regression problem over a continuous space, rather than a discrete (and more complex) categorical one. As discussed in Section \ref{sec:conclusion}, extension of this framework for LULC forecasting will be explored as future work, with related literature on categorical diffusion models corroborating the feasibility of such an extension \cite{hoogeboom2021argmax}. Inspired by its success in works such as Stable Diffusion \cite{rombach2022ldm} and MCVD \cite{voleti2022mcvd}, we adopted a modified SPADE scheme to condition our imperviousness forecasting diffusion model on two inputs: i) past imperviousness maps; and ii) past LULC maps.
Below we detail the dataset used for experimentation, the design of our diffusion-based framework, and describe a multiscale strategy for evaluating the predicted outputs.

\subsection{Dataset}\label{sec:dataset}
We use NLCD \cite{Wickham31122023, dewitz2023} provided by the U.S. Geological Survey (USGS) in association with the Multi-Resolution Land Characteristics (MRLC) Consortium. This data contains land cover and related information for the entire conterminous United States (CONUS), across nine epochs from 2001 to 2021 (2001, 2004, 2006, 2008, 2011, 2013, 2016, 2019, and 2021\footnote{during preparation of this manuscript, USGS has released a new version of NLCD ranging from 1985-2023 with annual timesteps \cite{usgs2024}. Experiments with this augmented dataset are targeted for future work}). It is based on spectral (i.e. Landsat bands), spatial, temporal, and ancillary data paired with land cover classification models, and is the most definitive source of LULC for the United States.

The NLCD data products provide raster information at a $30m/px$ resolution for each assessed year. Its urban imperviousness maps indicate impervious surfaces as a percentage of developed surface over every pixel, while its LULC maps follow a 16-class legend: \textit{Open Water, Perennial Ice/Snow, Developed (Open Space), Developed (Low Intensity), Developed (Medium Intensity), Developed (High Intensity), Barren Land (Rock/Sand/Clay), Deciduous Forest, Evergreen Forest, Mixed Forest, Shrub/Scrub, Grassland/Herbaceous, Pasture/Hay, Cultivated Crops, Woody Wetlands, Emergent Herbaceous Wetlands.} We use both these data products, from the nine epochs listed above, for training our diffusion models.

\subsection{Diffusion-model overview}\label{subsec:network-overview}


We largely follow \citet{dhariwal2021diffusion} for training our diffusion models, with modifications in the architecture to adapt our conditioning schemes.       As shown in Fig. \ref{fig:diffusion_diagrams}(a), training a diffusion model to learn from the data involves a forward and a reverse process. Mathematically, in the forward diffusion process a sample $x_0 \sim q(x_0)$ is corrupted between timesteps $t=0$ to $t=T$ with a Markovian noising process $q$ that gradually adds noise with the following kernel:  
\begin{equation}
    \label{eq:transition-kernel}
    q(x_t | x_{t-1}) = \mathcal{N}(x_t; \sqrt{1 - \beta_{t}} x_{t-1}, \beta_t\textbf{I})
\end{equation}
where $\beta_t$ is variance at step $t$. Under this formulation, $q(x_{1:T} | x_0) = \prod^T_{t=1}q(x_t|x_{t-1})$, which is tractable through the adoption of the normal distribution, but is computationally expensive as it requires applying the diffusion process several times. Instead, DDPM \cite{ho2020denoising} introduces a reparameterization trick: with $\alpha_t = 1 - \beta_t$ and $\Bar{\alpha_t} = \Pi^{t}_{s=0} \alpha_s$, then:
\begin{equation}
\begin{split}
    \label{eq:variance-fn}
    q(x_t | x_0) &= \mathcal{N}(x_t; \sqrt{\Bar{\alpha}} x_0, (1-\Bar{\alpha})\textbf{I})\\
    &= \sqrt{\Bar{\alpha}} x_0 + \epsilon\sqrt{1 - \Bar{\alpha}}, \epsilon \sim \mathcal{N}(0, \textbf{I})
\end{split} 
\end{equation}
where $1-\Bar{\alpha_t}$ is the variance of arbitrary timestep noise used to define noise schedule instead of $\beta_t$ \cite{dhariwal2021diffusion}. Since $\beta_t$, $\alpha_t$, $\bar{\alpha}_t$ can be precomputed at all timesteps, this formulation enables sampling $x_t$ at any arbitrary timestep, enabling a much simpler model training objective based on denoising.

\noindent\textbf{Cost Function}
For such reverse process, a parameterized conditional CNN is trained to recover the structure of the data by estimating the added noise (see Fig. \ref{fig:diffusion_diagrams}-b) at each node of the Markov chain. We train a conditional U-Net \cite{olaf2015unet} with a mean-squared error loss 
\begin{equation}
    \label{eq:mse}
    \left\lVert q_\theta(x_t, t) - q \right\rVert^2,
\end{equation}
where the $q$ denotes the true noise and $q_\theta$ the predicted noise at different steps of $x_t$ \cite{dhariwal2021diffusion}. 


\subsection{Conditioning data}\label{sec:condn-data}
In addition to conditioning the model on imperviousness maps from previous years, we lay the foundations to exploit additional sources of information. In particular, LULC labels are extremely relevant for imperviousness forecasting since change in imperviousness of a regions is directly correlated to change in the corresponding LULC category. For instance, areas such as \textit{water} and \textit{forest} are much less likely to experience imperviousness increase than categories such as \textit{pasture, developed (open-space, low, medium)}. Further, traditional methods such as CA-Markov rely on suitability matrices for LULC change forecasting by capturing the transition statistics between pairs of LULC categories at previous timestamps. We adapt this concept in the form of Imperviousness Likelihood Maps as described below.



\noindent\textbf{LULC-based Imperviousness Likelihood Maps:} We
compute imperviousness transition likelihood using transition statistics of LULC maps across successive timestamps. Algorithm \ref{alg:likelihood-algo} summarizes the steps to estimate such a likelihood map $LMap$. Let $C$ denote the number of LULC categories in the dataset. Given a pair of consecutive maps $lc_{t}$ and $lc_{t+1}$, first we construct a $C\times C$ transition matrix $T_{count}$ counting the number of pixel-level transitions from each category to another (i.e., their cross tabulation). 

\begin{algorithm}[t]
\footnotesize
\begin{algorithmic}
    \State \textbf{Inputs:} $lc_{t}, lc_{t+1} \in \mathbb{R}^{P\times P}$
    \State \textbf{Output:} $LMap_{t}$
    \State $T_{count} \gets crosstab(lc_{t}, lc_{t+1}, C)$ \Comment{$T_{count} \in \mathbb{R}^{CxC}$}
    \State \textit{Agg. into perv. and imperv. cols:}
    \State $T_{count}[:,0]  \gets sum\_cols(T_{count} , L_{perv})$ 
    \State $T_{count}[:,1] \gets weighted\_sum\_cols(T_{count}, L_{imperv}, W_{imperv})$
    \State $T_{prob.} \gets \frac{T_{count}}{row\_sum(T_{count})}$ \Comment{$T_{count} \in \mathbb{R}^{C\times2}$}
    \State $LMap_t \gets zeros(size={P\times P}$)
    \For{r in range($R$)}
    \For{c in range($C$)}
    \State $LMap_t[r,c] \gets T_{prob}[lc_t[r,c], 1]$ \Comment{lookup operation}
    \EndFor
    \EndFor
\end{algorithmic}

\caption{Imperviousness Likelihood Estimation}
\label{alg:likelihood-algo}
\end{algorithm}

\begin{table}[b]
    \caption{Imperviousness Levels and their LULC categories}
    \label{tab:imp_levels}
    \centering
    \begin{tabular}{l|l}
        \toprule
         Imperviousness Level & LULC Category \\
         \midrule
         $\leq 20\%$ & Developed (Open Space)  \\
         $20-49\%$ & Developed (Low Intensity)  \\
         $50-79\%$ & Developed (Medium Intensity) \\
         $\geq 80\%$ & Developed (High Intensity) \\
         \bottomrule
    \end{tabular}
\end{table}

By definition, the NLCD data associates non-zero imperviousness levels only to the four ``\textit{developed}'' categories, labeled according to their percentage of impervious surfaces as noted in Table \ref{tab:imp_levels}. Since we are interested in imperviousness forecasting, we opt to group these categories into \textit{pervious} ($C_{perv}$) and \textit{impervious} ($C_{imperv}$) subsets. To do so, we aggregate all the counts of the $C-4$ \textit{``no-development''} categories into a single column of ``fully-pervious'' counts. In turn, the four \textit{``developed''} categories are aggregated through a weighted sum, where the upper limits of impervious surface percentage are used as weights for the corresponding categories. This follows NLCD's design where by definition no decreases in imperviousness occur from one time step to the next.  

This reduces $T_{count}$ to a $C\times2$ matrix, which we finally convert into a $T_{prob}$ matrix by dividing each entry by its row sum. That is, each row in $T_{prob} \in \mathbb{R}_{C\times2}$ adds to $1$, representing the likelihood that each LULC category will transition into a pervious or impervious state. We can then generate an imperviousness transition likelihood map $\Lambda$ by using $T_{prob}$ as a lookup table: for a $(i,j) \in lc_t$ labeled with category $c_{ij}$, we assign $\Lambda(i,j)=T_{prob}(c_{ij}, 1)$, where column $1$ corresponds to imperviousness transition likelihood. Since two consecutive land cover maps are used to generate one likelihood map, $N-1$ likelihood maps are generated out of $N$ historical land cover maps. To overcome that, we generate an extra \textit{N-th} likelihood map by performing the lookup operation using $T_{prob}$ from the timestamp $t$ but based on land cover map $lc_{t+1}$.

\subsection{Conditioning architecture}\label{subsec:conditional-normalization}
%


Similar to MCVD \cite{voleti2022mcvd}, we adapt the concept of spatially-adaptive normalization \cite{park2019semantic} to enable conditioning the diffusion process on historical data. Normalization is a two-step process: i) mean normalization; ii) scaling by a $\gamma$ factor and a shifting by a $\beta$ factor\footnote{unrelated to the $\beta$ present in the diffusion model formulation Eq. \ref{eq:transition-kernel}}. As illustrated in Fig. \ref{fig:diffusion_diagrams}(d), SPADE \cite{park2019semantic} relies on a shallow CNN to learn $\gamma$ and $\beta$ from the conditioning inputs. The CNN has three layers: a simple \textit{Conv} layer, a \textit{ReLU}, then feeding two parallel \textit{Conv} layers for estimation of $\gamma$ and $\beta$. 





\noindent\textbf{Double Conditioning: } While SPADE and MCVD condition models on a single-modality data stream, we seek a framework for conditioning on multiple streams. Rather than adding a second $[Conv,ReLU,Conv]$ branch, we build on the intuition that past imperviousness $I_t$ and transition likelihood $\Lambda_t$ maps are spatially aligned and closely related, such that early-fusion can be pursued. Specifically, our conditioning architecture employs a shared $1\times1$ conv layer that takes as input a pair $[I_t, \Lambda_i]$ stacked along the channel dimension, and generates a single channel output. The weights of this $1\times1$ conv are shared across all $[I_t,L_t]$ pairs for $t=[0,N]$, with the outputs concatenated across the $N$ conditioning timestamps and passed as input to the \textit{[conv,ReLU,conv]} SPADE-like block to estimate $\gamma$ and $\beta$. In our implementation, we augment the architecture in \citet{dhariwal2021diffusion} by adding such conditioning of ($\gamma$,$\beta$) to the group normalization (GN) layers \cite{wu2018group} within each ResNet block composing the UNet trained for denoising. 

\subsection{Addressing data imbalance}
Since over $95\%$ of the United States corresponds to rural areas, most of the territory experienced little to no change in land cover across the years. This represents a major data imbalance, such that naively training models without any counterbalancing strategy leads to models that merely replicate data from previous years, without ever synthesizing any change. In addition to restricting all analyses to Metropolitan Statistical Areas (MSAs) as defined by the U.S. Census Bureau, we address this issue through a mechanism based on temporal clustering. First, for each $128\times128px$ patch we construct temporal signatures by computing the percentage of imperviousness change between each pair of subsequent years composing the NLCD dataset in the interval of 2001-2019. Then, we employ dynamic time warping (DTW) \cite{berndt1994usingDTW} to group the patches (regions) into $5$ clusters. 
\begin{figure}[h]
    \centering
    \includegraphics[width=\linewidth]{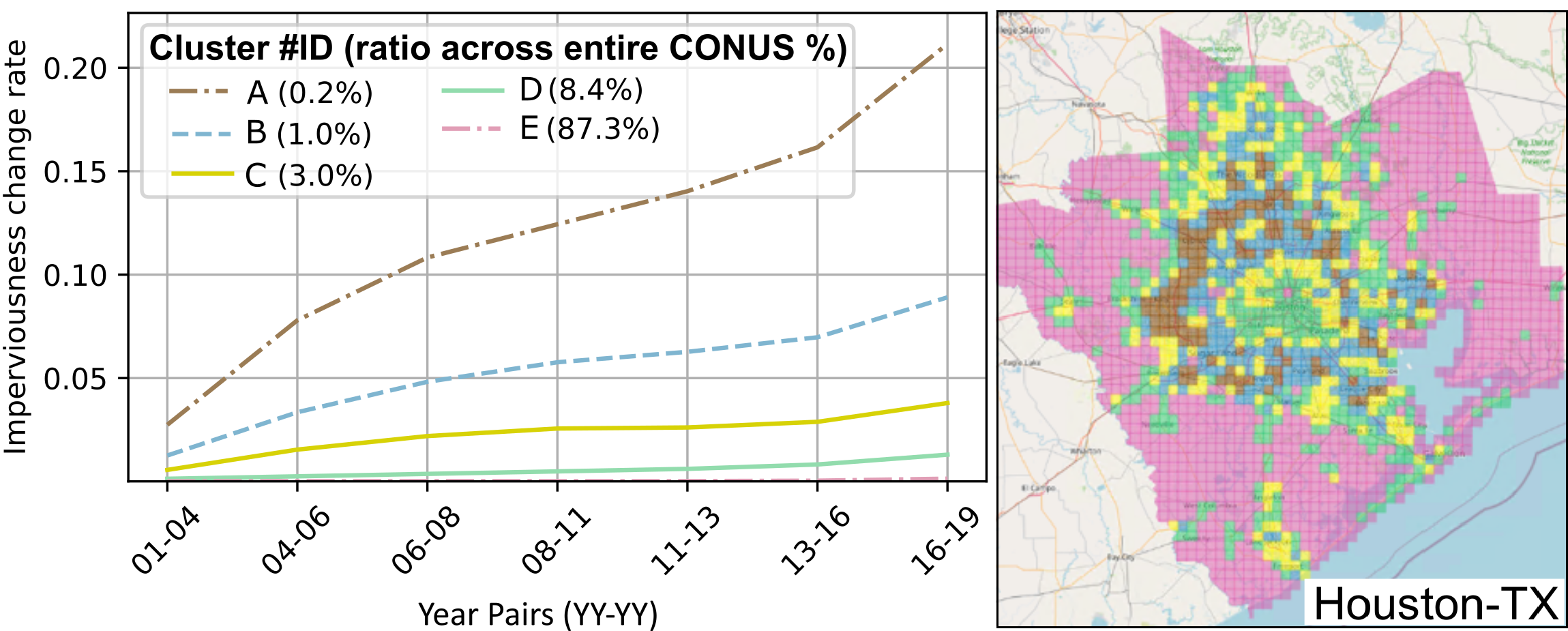}
    \caption{Summary of clusters identified through DTW \cite{berndt1994usingDTW} based on imperviousness changes across consecutive pairs of years in NLCD. \textit{Right:} illustration of spatial distribution of DTW clusters for Houston-TX (MSA)}
    \label{fig:dtw}
\end{figure}

Figure \ref{fig:dtw} illustrates the average centroid signature for each cluster, together with the percentage of patches assigned to each, and an illustration of their spatial distribution for the Houston (Texas) area. It shows how heavily imbalanced the data is even when restricted to MSAs: over $87\%$ of the data (cluster \#E) corresponds to areas with little to none imperviousness change as they did not experience significant urbanization; additional $>8\%$ (cluster \#D) also experience little change, but instead mostly because they correspond to heavily urbanized areas that had already reached nearly $100\%$ imperviousness. Based on the cluster composition ratios, we implement a reverse weighting scheme to expose the model to samples from each cluster nearly equally during training.

\begin{figure}[h]
    \centering
    \includegraphics[width=\linewidth]{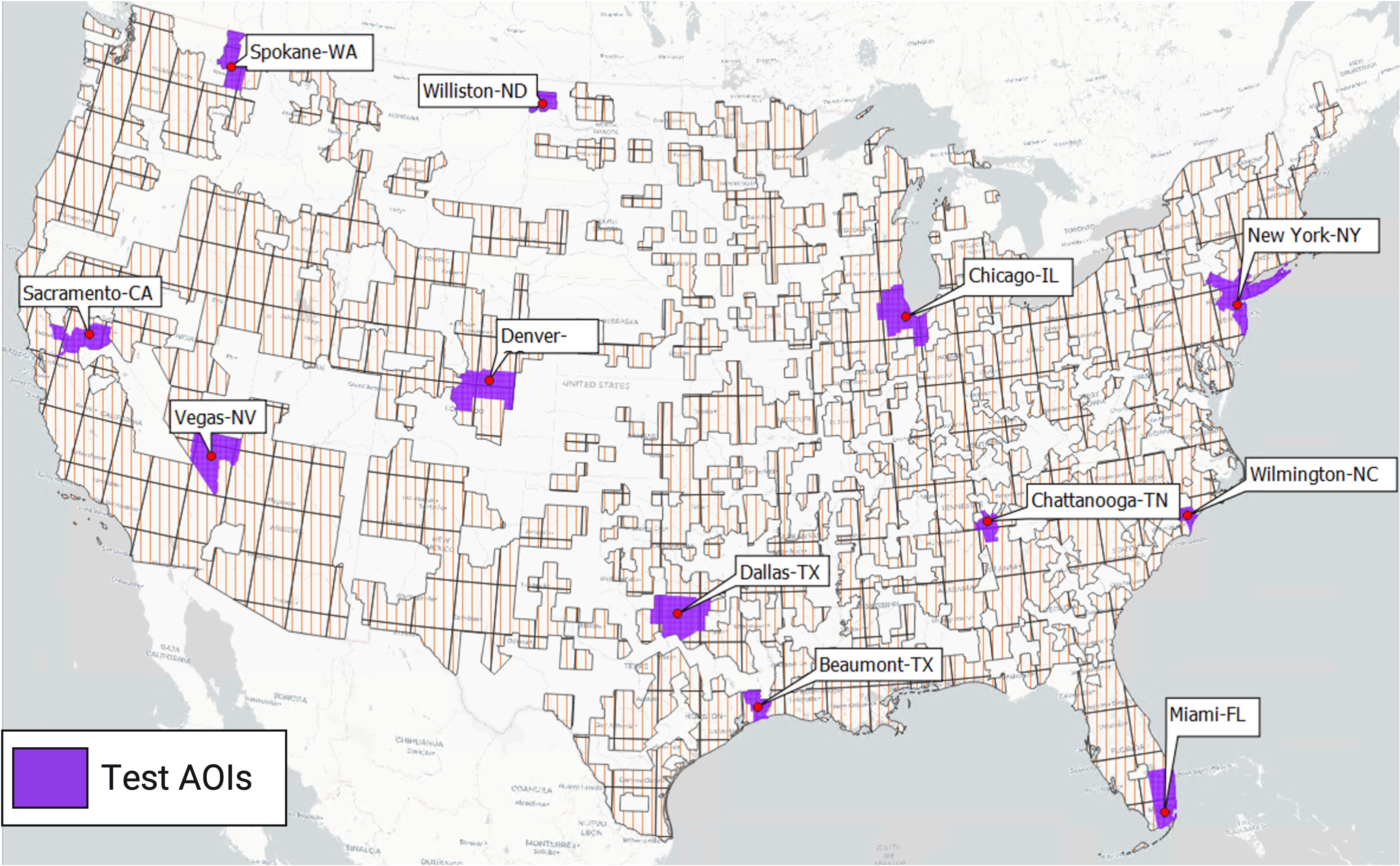}
    \caption{Spatial distribution of the MSAs used for training and evaluation of our imperviousness forecasting model. For testing, data from $2021$ is held-out during training.}
    \label{fig:splits}
\end{figure}

\subsection{Multi-scale spatial evaluation}\label{subsec:spatial-evaluation}
The heavy imbalance of most areas not experiencing any LULC change over years (or even decades) means land cover change forecasting is a problem for which even a simple baseline ``null'' model that assumes no change between years (i.e., pure persistence) can yield high accuracy at pixel-level \cite{pontius2004null, pontius2005comparison, van2011revisiting}. In addition to the imbalance problem, metrics based solely on pixel-by-pixel ``hit or miss'' lack spatial awareness: e.g., an incorrect change prediction (``false alarm'') for a pixel neighboring a pixel where change did occur (``true positive'') is penalized in the same way as if no change had occurred at all in its broader neighborhood, which is intrinsically incorrect for such a geospatial task.

\begin{figure}[t]
\begin{center}
   \includegraphics[width=\linewidth]{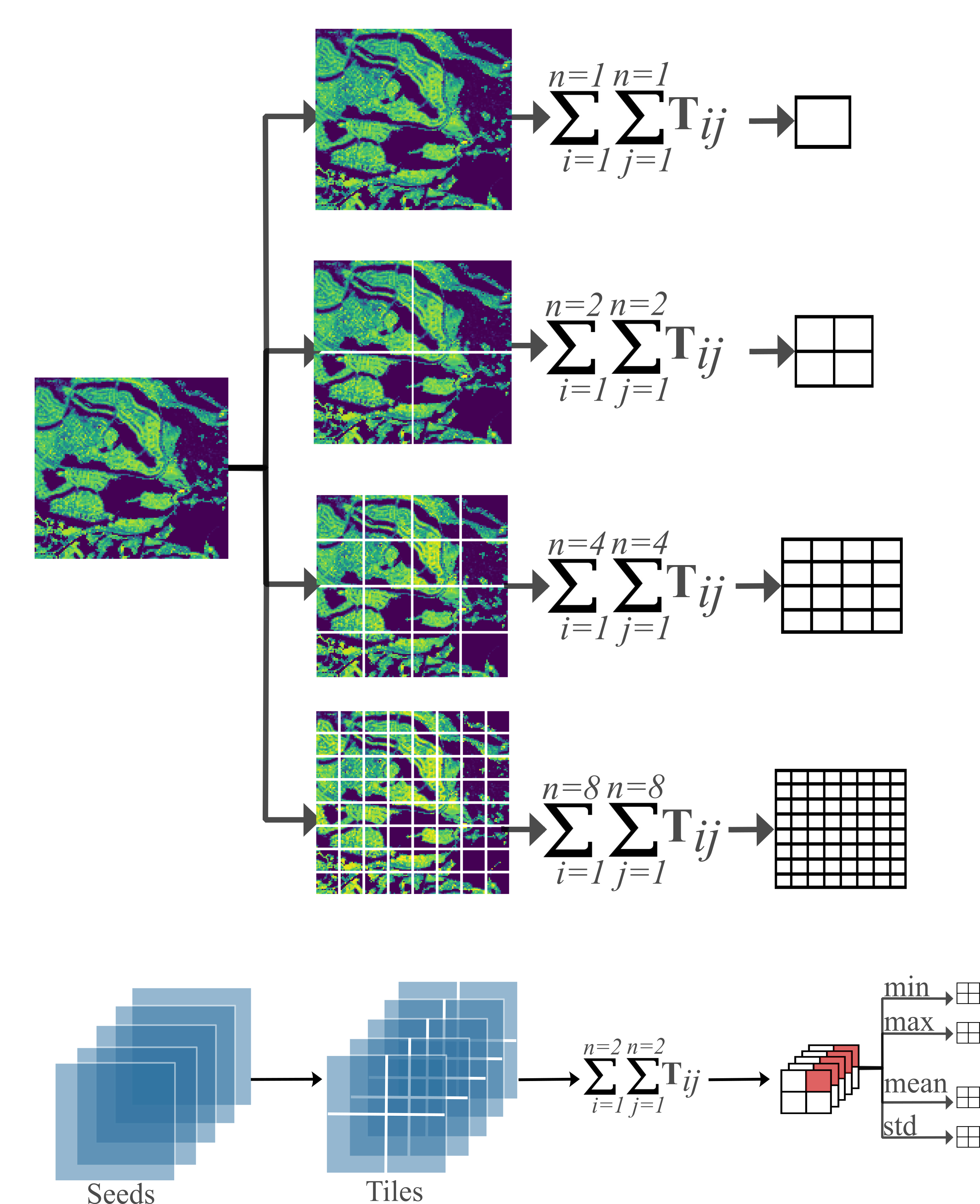}
    \caption{Illustration of the multi-scale evaluation scheme adopted in this work. \textit{Top:} the spatial aggregation process, where pixel-level predictions of imperviousness are aggregated into cells of varied sizes. \textit{Bottom:} the statistics computed across the multiple predictions provided by our diffusion model, each with a different random initialization seed. }
    \label{fig:spatial-evaluation}
\end{center}
\end{figure}

In \cite{pontius2005comparison}, Pontius et al. introduce the notion of ``null resolution'' for assessing land change models: the resolution at which the accuracy of the predictive model equals the accuracy of a null model. Here, models with a finer null resolution reflect higher accuracy and precision in their predictions. We adapt this concept for evaluation of imperviousness forecasting models by computing the mean absolute error (MAE) at varied spatially aggregated cells. Figure \ref{fig:spatial-evaluation} illustrates this process: pixel-level predictions are gradually coarsened through a process of spatial aggregation within cells of increasingly large sizes. 

\section{Experimental Setup} \label{sec:experiments}
\noindent\textbf{Data splits} Across all experiments, we adopt patch sizes of $128\times128 px$, equivalent to $\approx 3.8\times3.8 km^2$, resulting in 416,475 tiles across the CONUS. Figure \ref{fig:splits} illustrates the MSAs, with model training taking place using NLCD data from years 2001-2019. We holdout 2021 for testing, which is conducted for the 12 areas of interest (AOIs) selected across the CONUS: Wilmington-NC, New York-NY, Dallas-TX, Las Vegas-NV, Sacramento-CA, Williston-ND, Miami-FL, Chicago-IL, Spokane-WA, Denver-CO, Beaumont-TX, \& Chattanooga-TN. These areas were selected such that geographical diversity as well as diversity in terms of population density and pace of development are present.

\vspace{10pt}\noindent\textbf{Model training} We train our diffusion models for 10-year forecasting based on $N=3$ timestamps of past imperviousness and transition likelihood maps. From the available NLCD years [2001, 2004, 2006, 2008, 2011, 2013, 2016, 2019], for each batch our training dataloader selects a target year from [2016, 2019], with the corresponding $N=3$ timestamps $\geq10$ years older used for conditioning (e.g., [2004, 2006, 2008] as conditions to reconstruct 2019).
In addition to a \textit{national} model trained on data sampled from all patches across the CONUS, we also train four individual models on each of the Clusters A, B, C, and D to investigate if an ensemble of cluster-specialist models is advantageous over a single model trained on more diverse data. Since samples associated with Cluster E go through little to no change over the years, instead of training a model on this cluster we assume persistance (i.e., no change) for such areas. We refer to these five models as \textsc{GeoDiff-National}, \textsc{GeoDiff-ClusterA, GeoDiff-ClusterB, GeoDiff-ClusterC} and \textsc{GeoDiff-ClusterD}, respectively. All models are trained for the equivalent of $20k$ iterations with a batch size of $512$, learning rate of $3e-4$, exponential moving average rate of $0.99$, a linear noise scheduler, and no weight decay. 




 \vspace{10pt}\noindent\textbf{Sampling and evaluation} For sampling, we generate $5$ random seeds for each validation tile, whose statistics we aggregate as described in Section \ref{subsec:spatial-evaluation} and Figure \ref{fig:spatial-evaluation}. More specifically, we aggregate pixel-level predictions and reference ground truth across squared cells of size $S\times S$, with $S=[4, 8, 16, 32, 64, 128]$ pixels. With a base resolution of $30 m/px$ this corresponds to aggregations into $[120m, 240m, 480m, 960m, 1920m, 3840m]^2$ cells, respectively. At each scale we compute the MAE between the imperviousness predictions by our diffusion model for 2021, and the reference NLCD data for 2021. Across all scales, we compare this with the MAE of a corresponding null model that assumes no change between 2011 and 2021 to enable  estimating the \textit{null resolution} of our diffusion model\footnote{to estimate the null resolution for each plot, we apply bicubic interpolation to the diffusion-based MAE curve and use the Brent’s method (\textit{scipy.optimize.brentq}) to find the root of its difference function to the null model's MAE}. At resolutions finer than the \textit{null resolution}, the diffusion model is less accurate than the null model, while at resolutions coarser than the null resolution, the diffusion model is more accurate than the null model \cite{pontius2005comparison}. We sample our models using DDIM \cite{song2021denoising} with 500 steps to achieve fast inference time. In average, it takes about $1.5 hrs$ to sample approximately $2000$ tiles using two NVIDIA A100 GPUs.

\section{Results and Discussion}\label{sec:results}

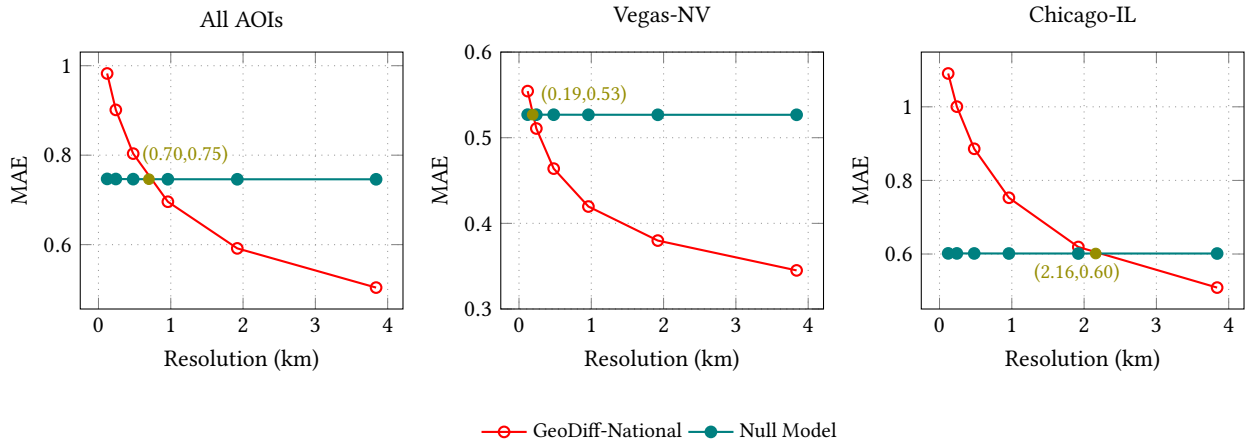
\begin{figure*}[t]
    \centering
    \begin{tikzpicture}
    \begin{groupplot}[
        group style={
            group size=3 by 1,
            horizontal sep=1.3cm,
        },
        width=0.33\textwidth,
        height=5cm,
        xlabel={Resolution (km)},
        ylabel={MAE},
        title style={align=center},
        grid=major,  
        grid style={dotted,gray!80},
        legend style={at={(1.8,-0.4)}, anchor=north, legend columns=-1, draw=none},
    ]

    \nextgroupplot[title={All AOIs}]
    \addplot[color=red, mark=o, thick] coordinates {(0.120,0.982636) (0.240,0.901225) (0.480,0.803475)(0.960,0.695939) (1.920,0.591636) (3.840,0.503988)};
    \addlegendentry{\small GeoDiff-National}
    \addplot[color=teal, mark=*, thick] coordinates {(0.120,0.746934) (0.240,0.746716) (0.480,0.746416) (0.960,0.746184) (1.920,0.746098) (3.840,0.746069)};
    \addlegendentry{\small Null Model}
    \addplot[only marks, mark=*, color=olive] coordinates {(0.6980350207767063,0.74631073)};
    \node at (axis cs:1.200,0.8) {\textcolor{olive}{\small(0.70,0.75)}};

    \nextgroupplot[ymin=0.3, ymax=0.6, title={Vegas-NV}]
    \addplot[color=red, mark=o, thick] coordinates {(0.120,0.554409) (0.240,0.510753) (0.480,0.463902) (0.960,0.419637) (1.920,0.379884) (3.840,0.345062)};
    \addplot[color=teal, mark=*, thick] coordinates {(0.120,0.526932) (0.240,0.526892) (0.480,0.526857) (0.960,0.526798) (1.920,0.526760) (3.840,0.526760)};
    \addplot[only marks, mark=*, color=olive] coordinates {(0.18942,0.52690879)};
    \node at (axis cs:0.9,0.55) {\textcolor{olive}{\small(0.19,0.53)}};



    \nextgroupplot[title={Chicago-IL}]
    \addplot[color=red, mark=o, thick] coordinates {(0.120,1.090583) (0.240,1.000458) (0.480,0.885915) (0.960,0.752842) (1.920,0.619066) (3.840,0.508764)};
    \addplot[color=teal, mark=*, thick] coordinates {(0.120,0.601712) (0.240,0.601656) (0.480,0.601536) (0.960,0.601412) (1.920,0.601355) (3.840,0.601355)};
    \addplot[only marks, mark=*, color=olive] coordinates {(2.16058,0.60135537)};
    \node at (axis cs:1.900,0.55) {\textcolor{olive}{\small(2.16,0.60)}};

    \end{groupplot}
    \end{tikzpicture}
    \caption{MAE vs Resolution of GeoDiff-National and a Null model across (a) all the AOIs, (b) Vegas-NV, and (c) Chicago-IL. A null resolution \cite{pontius2005comparison} of $x$ m  corresponds to an area of $x^2$ m$^2$ on land.}
    \label{fig:mae_plots}
\end{figure*}

\begin{table*}[h]
  \caption{Validation AOIs, their distribution across all clusters, and the Null Resolution by GeoDiff-National}
  \label{tab:aoi-model}
  \begin{tabular}{lcccccccccc}
    \toprule
    
    \multirow{2}{*}{AOI} & \multicolumn{5}{c}{Cluster Distribution ($\%$)} & 
  
    \multirowcell{2}{Mean Change \\ 2001-2011} &
    \multirowcell{2}{Mean Change \\ 2011-2021} &  
    \multirowcell{2}{STD Change \\ 2011-2021} &
      \multirowcell{2}{Null \\ Resolution (km)} & 
    \multirowcell{2}{MAE \\ @ NR} \\
    
    \cline{2-6}
        &  A & B & C & D & E &  \\
    \midrule   
    Vegas-NV & 2.93 &  3.08 &  3.23 &  4.40 & 86.35 & 0.34 & 0.21 & 3.57 & 0.19 &	0.53 \\
    Williston-ND & 0.60 &  2.38  & 3.57 & 23.21 & 70.24  & 0.07 & 0.39 & 4.77 & 0.37 & 0.74 \\
    Miami-FL & 1.37 &  4.28 &  8.73 & 13.70 & 71.92 & 0.56 & 0.38 & 4.83 & 0.44 & 0.53 \\
    Wilmington-NC & 0.00 &  1.64  & 9.21 & 13.82 & 75.33 & 0.20 & 0.22 &	3.51 & 0.44 & 0.46 \\
    Dallas-TX & 5.01 & 12.95 & 20.53 & 28.16 & 33.35 & 1.72 & 1.83  & 10.74 & 0.46 & 2.21 \\
    Sacramento-CA & 1.71  & 3.33  & 5.65 &  9.78 & 79.54 & 0.40 & 0.24 & 3.86 & 0.47 & 0.42 \\
    Denver-CO & 1.74 &  4.38  & 6.03 &  8.10 & 79.75  & 0.29 & 0.22   & 3.78 & 0.49 & 0.54\\
    Chattanooga-TN & 0.65  & 1.63 &  5.88 & 18.30 & 73.53 & 0.29 & 0.15 & 2.84 & 0.92 & 0.27 \\ 
    Spokane-WA & 0.00  & 1.12 &  1.69 &  5.48 & 91.71 & 0.13 & 0.09 & 2.24 & 1.09 & 0.13 \\ 
    Beaumont-TX & 0.17 &  1.01 &  5.04 & 16.47 & 77.31 & 0.22 & 0.17 & 3.23 & 1.50 & 0.28 \\ 
    New York-NY & 0.00  & 2.02 & 10.27 & 43.99 & 43.72 & 0.14 & 0.09 & 2.30 & 1.66 & 0.47 \\   
    Chicago-IL & 1.36 &  9.48 & 15.73 & 26.87 & 46.55  & 0.76 & 0.30 & 4.32 & 2.16 & 0.60 \\ 
    \bottomrule
\end{tabular}
\end{table*}

Figure \ref{fig:mae_plots} shows plots of MAE vs aggregation resolution for our proposed diffusion model and the corresponding null model. Overall, \textsc{GeoDiff-National} yields a sub-kilometer null resolution, around $0.7 km\times0.7 km$ across all 12 AOIs. This is a stark contrast to the city/region-specific models composing most of the related literature. 

Table \ref{tab:aoi-model} summarizes model performance in terms of null resolutions across each of the 12 validation AOIs. As expected, for regions with more substantial ground-truth changes our model starts outperforming a null baseline (which assumes pure persistence) at finer resolutions. High-change regions like\textit{ Denver, Miami, Vegas, Wilmington}, and \textit{Williston} all have null resolution $<0.5km\times0.5km$, with \textit{Vegas} showing as fine as $<0.2km\times0.2km$. In contrast, for low-change regions like \textit{New York, Spokane}, and \textit{Beaumont} it takes more aggregation (i.e. a coarser null resolution) for the model to beat the persistency baseline. 

Two outliers are noteworthy: while \textit{Dallas} has by far the highest rates of change (1.72 \& 1.83 for 2001-'11 \& 2011-'21, respectively), resulting in one of the finest null resolutions, its high percentages of Cluster C and D samples indicates several areas of low development on which our model seems to perform less well (more below). This is also evident from the highest STD of change captured from the ground truth analysis. On the other hand, the poor performance for \textit{Chicago} is better understood by comparing the statistics of imperviousness change between 2001-2011 (time period for model training and conditioning) as compared to 2011-2021 (time period for model prediction). While it presented the second highest pace of growth between 2001-2011, this dropped by more than 60\% in terms of mean increase in imperviousness. A similar shift in development pace is observed for \textit{New York}, for which the second coarsest null resolution is obtained.



Overall, these patterns corroborate how our model successfully captures spatial and temporal patterns of development from the conditioning data spanning 2006, 2008, \& 2011, with such a data-driven forecasting mechanism showing strong predictive abilities across several AOIs. However, the model implicitly learns an assumption of similar pace of development taking place in the future, which does not necessarily hold as land cover change is heavily dependent on factors such as human decision making. The time-span adopted for this experimentation is a clear example, as the 2008 financial crisis highly affected the pace of development over the subsequent decade.

\noindent\textbf{Cluster-by-cluster analysis} 
We also show the results of \textsc{GeoDiff-National} evaluated on each specific cluster in Table \ref{tab:clstr-model}. \textsc{GeoDiff-National}'s null resolution is again the finest for samples characterized by highest change (i.e. \textit{Cluster A}). This implies that the model is sensitive towards areas undergoing rapid change. On the other hand, the model shows a coarser null resolution for areas going through lesser change (such as samples from \textit{Cluster C} and \textit{Cluster D}). This suggests the model has a tendency to over-predict change in areas that experience slower development.

Table \ref{tab:clstr-model} also compares \textsc{GeoDiff-National} with the \textsc{GeoDiff} architecture trained on individual clusters instead. We see that these cluster-specific models can give a better null resolution for their respective clusters, with an ensemble of such models further improving the performance provided by the \textsc{GeoDiff-National} trained at the CONUS scale. This represents a trade-off, as training multiple models require investing in more compute. 

Further, Figure \ref{fig:mae_plots_clstrmodels} (Appendix) shows the MAE plots of each of the cluster-specific models evaluated across all their respective AOIs. The MAE of a null model decreases from \textit{Cluster A} to \textit{Cluster D}. While the diffusion model beats the null model by a large margin for fast-changing regions to achieve a fine null resolution, it becomes harder to beat a null model for slow-changing regions. 
This observation reinforces how models adopting a conservative approach of predicting small quantities of change tend to have lesser overall errors for such an imbalanced problem, but on turn have higher errors when the assumption does not hold \cite{pontius2005comparison}.

\begin{figure*}[ht]
    \centering
    \includegraphics[width=\linewidth]{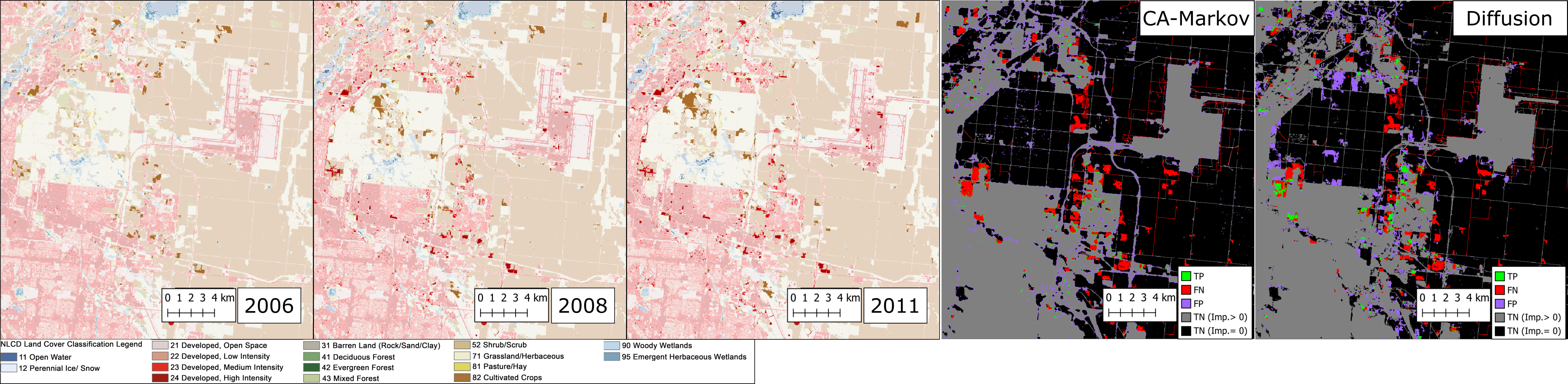}
    \caption{Qualitative comparison between 2021 imperviousness change forecasts from a CA-Markov model (conditioned on 2001-2011) and our \textsc{GeoDiff-National} model (conditioned on [2006, 2008, 2011] for an area in Denver-CO. \textit{Left:} Three left-most panels show LULC maps across conditioning years, with areas that underwent change highlighted. \textit{Right:} Predictions from CA-Markov and our model, compared to NLCD's 2021 map. True Positives (TP), False Positives (FP), True Negatives (TN) are color-coded based on binary comparison of change (i.e., imperviousness change $>0$).}
    \label{fig:denverQuali}
\end{figure*}

\begin{table}[h]
  \caption{Null resolution of GeoDiff-National and the cluster-specific models (GeoDiff-Cluster*) across different clusters.}
  \label{tab:clstr-model}
  \begin{tabular}{ccc}
    \toprule
    Cluster & GeoDiff-Cluster* & GeoDiff-National \\ 
    \midrule
    Cluster A & < 0.1 km &  0.2 km \\ 
    Cluster B  & < 0.1 km & 0.5 km \\ 
    Cluster C & < 0.1 km & 0.9 km  \\ 
    Cluster D & 0.9 km  &  1.7 km \\ 
    \midrule
    All Clusters & < 0.1 km &  0.7 km \\ 
  \bottomrule
\end{tabular}
\end{table}

\subsection{Qualitative Case Study}
We also conducted a qualitative analysis to provide further insights particularly on the spatial patterns of change predicted by our proposed method. Specifically, we perform inference using GeoDiff-National over the Denver AOI, comparing its predictions to a CA-Markov approach to assess how our diffusion-based approach stands against this traditional and popular approach of urban growth modeling. 

\noindent\textbf{CA-Markov baseline} Following \citet{verburg2002modeling}, we first derive a transition probability matrix based on LULC transitions between $2001$ and $2011$ for a subset of 8 NLCD land cover classes in the Denver AOI: Water, Developed, Barren, Forest, Shrubland, Herbaceous, Cultivated, and Wetlands. We then use this matrix to project total LULC change by area for $2021$. LULC is then allocated to each grid cell based on suitability grids for each LULC, which are generated based on both the transition probabilities and a CA filter that incorporates neighborhood effects. The allocation algorithm in \citet{verburg2002modeling} is essentially a greedy search: i) cell-level allocations are initially defined on their highest LULC probability; ii) the overall LULC allocation in the AOI is then compared to the projected levels, highlighting which LULC types are over-/under-allocated; iii) the difference between the total and projected levels is minimized by dynamically updating the probabilities of each LULC (increase probability of under-allocated types, decrease for over-allocated types). This cycle is repeated until the land allocation sums to the total expected LULC change for each type. At the end, we obtain binary maps of imperviousness change by thresholding on the corresponding Developed category.

Figure \ref{fig:denverQuali} shows the qualitative differences between predictions provided by this CA-Markov baseline and our diffusion-based model. To support the analysis, we also provide in Table \ref{tab:confusion_metrics} the True Positives (TP), False Positives (FP), False Negatives (FN), Precision, Recall, and F-1 score by the two methods, as compared to the actual change documented in NLCD between years $2011-2021$. Since such a baseline cannot directly offer outputs in a continuous range (i.e., only classification instead of regression), only such a binary quantitative analysis of change/no-change is possible. 

Overall, the diffusion-based approach presents significantly higher recall and precision rates. From a visual analysis of spatial patterns in Figure \ref{fig:denverQuali}, it is noticeable how the CA-Markov approach heavily relies on spatial connectivity, while GeoDiff synthesizes larger clusters of change. This is evident in the form of larger TP areas (in particular, ``infilling'' within areas of prior high-development), as well as FPs: while a large volume of FPs for CA-Markov are at borders of previously-developed areas (such as roads), GeoDiff synthesizes new clusters. Moreover, a comparison with the conditioning maps shows how GeoDiff synthesizes changes mostly in areas that have undergone development in previous years -- for instance, in the vicinity of the center-left area where there was an increase on Cultivated Crops (brown), and the bottom-center area where there was an increase from Low to High Intensity development (red).

\begin{table}[ht]
    \centering
    \caption{Confusion Matrix Metrics for CA and Diffusion}
    \setlength{\tabcolsep}{4.5pt}
    \begin{tabular}{lcccccc}
        \toprule
        & TP & FP & FN & Recall & Precision & F1 \\
        \midrule
        CA & 13894 & 262870 & 111589 & 11.07 & 5.02 & 6.91 \\
        Diffusion & 21163 & 138495 & 104320 & 16.87 & 13.26 & 14.84 \\
        \bottomrule
    \end{tabular}
    \label{tab:confusion_metrics}
\end{table}


\section{Conclusion}\label{sec:conclusion}

We proposed a new data-driven paradigm for land-cover change forecasting exploiting modern generative AI, showcasing how diffusion models can be used to implicitly identify patterns in urban land cover change and thus to forecast plausible future land-cover change outcomes. Specifically, we described experiments on forecasting imperviousness change by conditioning our model on historical timesteps as well as auxiliary information in the form of LULC transition likelihood maps. Results show that a single model trained at national scale under the proposed paradigm is capable of providing sub-kilometer null resolutions across multiple regions, a stark contrast to conventional region-specific models. Moreover, inference for such regions can be achieved in a few hours, as compared to days taken by CA-Markov like models. We also show how spatio-temporal patterns captured by our data-driven model go beyond assumptions of spatial connectivity while showing visually plausible layouts. 

\noindent\textbf{Limitations and Future Work} Our models make forecasts based on how land-cover has changed in the past. However in reality, land-use patterns change due to a variety of human and ancillary drivers such as population dynamics, land use policies, local terrain etc. which might lead to change that hadn't been observed in the past. Hence, there is a scope to condition and constrain our model with such driver variables to not only improve forecasting accuracy, but to also derive insights about the relationship between these variables and their affect on land-cover. While our evaluation scheme based on multi-scale aggregation enables capturing some patterns of spatial distribution, development of evaluation metrics that further capture concepts such as spatial heterogeneity and consistency, as well as prior knowledge, could enable a more thorough evaluation towards better forecasting models \cite{brelsford2020urban}. Finally, we plan to extend the framework for categorical LULC forecasting by exploring approaches for diffusion-based discrete and categorical data synthesis, such as combining argmax flows with multinomial diffusion \cite{hoogeboom2021argmax}. 

\begin{acks}
 This manuscript has been authored by UT-Battelle, LLC, under contract DE-AC05-00OR22725 with the US Department of Energy (DOE). The US government retains and the publisher, by accepting the article for publication, acknowledges that the US government retains a nonexclusive, paid-up, irrevocable, worldwide license to publish or reproduce the published form of this manuscript, or allow others to do so, for US government purposes. DOE will provide public access to these results of federally sponsored research in accordance with the DOE Public Access Plan (http://energy.gov/downloads/doe-public-access-plan). This study is supported by the U.S. Department of Energy, Office of Science, Biological and Environmental Research Program under Award Number DE-SC0023216.
\end{acks}

\bibliographystyle{ACM-Reference-Format}
\bibliography{main}

\appendix

\section{Appendix}\label{sec:appendix}

\begin{figure}[h]
\centering
\begin{tikzpicture}
\begin{groupplot}[
    group style={
        group name=my plots,
        group size=2 by 2, 
        horizontal sep=1.3cm,
        vertical sep=1.7cm,
    },
    width=0.48\linewidth, 
    xlabel={Resolution (km)},
    ylabel={MAE},
    legend style={font=\tiny},
    title style={align=center},
    legend style={at={(-0.2,-0.5)}, anchor=north, legend columns=-1, draw=none},
    grid=major,  
    grid style={dotted,gray!80},
]

\nextgroupplot[title={GeoDiff-ClusterA}]
\addplot[color=red, mark=o, thick] coordinates {(0.120,4.092797) (0.240,3.876438) (0.480,3.672370)(0.960,3.503664) (1.920,3.412397) (3.840,3.356301)};
\addplot[color=teal, mark=*, thick] coordinates {(0.120,8.708861) (0.240,8.707932) (0.480,8.706818) (0.960,8.706244) (1.920,8.706244) (3.840,8.706244)};

\nextgroupplot[title={GeoDiff-ClusterB}]
\addplot[color=red, mark=o, thick] coordinates {(0.120,3.038121) (0.240,2.864801) (0.480,2.667733) (0.960,2.467820) (1.920,2.299842) (3.840,2.198544)};
\addplot[color=teal, mark=*, thick] coordinates {(0.120,4.521010) (0.240,4.519878) (0.480,4.518454)(0.960,4.517768) (1.920,4.517740) (3.840,4.517741)};

\nextgroupplot[title={GeoDiff-ClusterC}]
\addplot[color=red, mark=o, thick] coordinates {(0.120,1.765655) (0.240,1.615931) (0.480,1.448164)(0.960,1.281078) (1.920,1.122400) (3.840,0.978358)};
\addplot[color=teal, mark=*, thick] coordinates {(0.120,1.871831) (0.240,1.871079) (0.480,1.870122) (0.960,1.869410) (1.920,1.869252) (3.840,1.869252)};

\nextgroupplot[title={GeoDiff-ClusterD}]
\addplot[color=red, mark=o, thick] coordinates {(0.120,1.086550) (0.240,0.969504) (0.480,0.822072)(0.960,0.672612) (1.920,0.517318) (3.840,0.376306)};
\addlegendentry{\small GeoDiff-Cluster*}
\addplot[color=teal, mark=*, thick] coordinates {(0.120,0.690176) (0.240,0.689836) (0.480,0.689285) (0.960,0.688771) (1.920,0.688491) (3.840,0.688384)};

\addlegendentry{\small Null Model}

\end{groupplot}
\end{tikzpicture}
\caption{MAE vs Resolution of the four cluster-specific models across all their AOIs}
\label{fig:mae_plots_clstrmodels}
\end{figure}
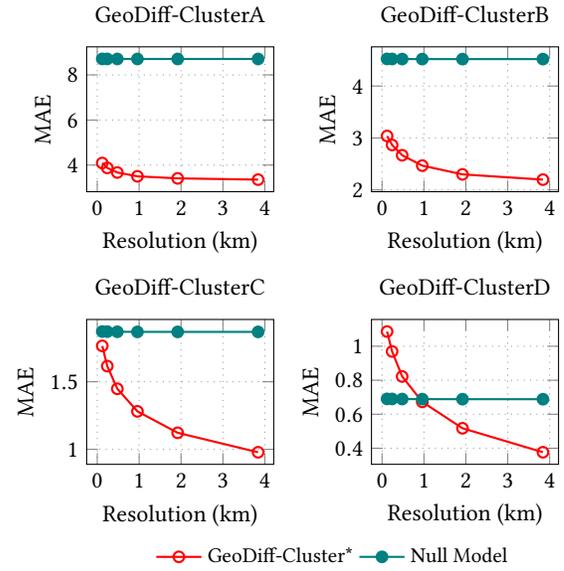

Cluster-specific models \textsc{GeoDiff-ClusterA}, \textsc{GeoDiff-ClusterB}, \textsc{GeoDiff-ClusterC}, and \textsc{GeoDiff-ClusterD} show decrease in MAE but their respective null resolution becomes coarser (Figure \ref{fig:mae_plots_clstrmodels}).



\end{document}